\documentclass[journal,twoside,web]{ieeetran}

\usepackage{cite}
\usepackage{amsmath,amssymb,amsfonts}
\usepackage{algorithmic}

\usepackage{amsmath}
\usepackage{booktabs}
\usepackage{booktabs}
\usepackage{multirow}

\usepackage{hyperref}

\usepackage{graphicx}
\usepackage{textcomp}
\usepackage{amssymb}
\usepackage{pifont}
\usepackage{bm}

\usepackage{caption}
\usepackage{array} 

\newcommand{\hl}[1]{#1}

\usepackage{fancyhdr}

\def\@BTrule[#1]{%
  \ifx\longtable\undefined
    \let\@BTswitch\@BTnormal
  \else\ifx\hline\LT@hline
    \nobreak
    \let\@BTswitch\@BLTrule
  \else
     \let\@BTswitch\@BTnormal
  \fi\fi
  \global\@thisrulewidth=#1\relax
  \ifnum\@thisruleclass=\tw@\vskip\@aboverulesep\else
  \ifnum\@lastruleclass=\z@\vskip\@aboverulesep\else
  \ifnum\@lastruleclass=\@ne\vskip\doublerulesep\fi\fi\fi
  \@BTswitch}
\makeatother

\begin{document}
\title{NeeCo: Image Synthesis of Novel Instrument States Based on Dynamic and Deformable 3D Gaussian Reconstruction}

\author{Tianle Zeng, Junlei Hu, Gerardo Loza Galindo, Sharib Ali, Duygu Sarikaya, Pietro Valdastri 
, \ and Dominic Jones*
\thanks{This paper was submitted for review on 9th August 2025}
\thanks{T. Zeng,  J. Hu, P. Valdastri, and D. Jones are with the STORM Lab UK, School of Electronic and Electrical Engineering, University of Leeds, LS2 9JT, Leeds, UK. (corresponding author e-mail: D.P.Jones@leeds.ac.uk).
G. L. Galindo, S. Ali, D. Sarikaya are with the STORM Lab UK, and School of Computer Science, University of Leeds, LS2 9JT, Leeds, UK.}}

\pagestyle{fancy} 
\fancyhead{} 
\renewcommand{\headrulewidth}{0pt}  
\fancyhead[L]{This work has been submitted to the IEEE for possible publication. Copyright may be transferred without notice, after which \\ this version may no longer be accessible.} 

\maketitle
\thispagestyle{fancy}
\begin{abstract}
Computer vision-based technologies significantly enhance surgical automation by advancing tool tracking, detection, and localization. However, Current data-driven approaches are data-voracious, requiring large, high-quality labeled image datasets, which limits their application in surgical data science. Our Work introduces a novel dynamic Gaussian Splatting technique to address the data scarcity in surgical image datasets. We propose a dynamic Gaussian model to represent dynamic surgical scenes, enabling the rendering of surgical instruments from unseen viewpoints and deformations with real tissue backgrounds. We utilize a dynamic training adjustment strategy to address challenges posed by poorly calibrated camera poses from real-world scenarios. Additionally, we propose a method based on dynamic Gaussians for automatically generating annotations for our synthetic data. For evaluation, we constructed a new dataset featuring seven scenes with 14,000 frames of tool and camera motion and tool jaw articulation, with a background of an ex-vivo porcine model. Using this dataset, we synthetically replicate the scene deformation from the ground truth data, allowing direct comparisons of synthetic image quality. Experimental results illustrate that our method generates photo-realistic labeled image datasets with the highest values in Peak-Signal-to-Noise Ratio (29.87). We further evaluate the performance of medical-specific neural networks trained on real and synthetic images using an unseen real-world image dataset. Our results show that the performance of models trained on synthetic images generated by the proposed method outperforms those trained with state-of-the-art standard data augmentation by 10\%, leading to an overall improvement in model performances by nearly 15\%.

\end{abstract}

\begin{IEEEkeywords}
Surgical Data Science, Surgical AI, Data generation, 3D Gaussian splatting, Laparoscopy.
\end{IEEEkeywords}

\section{Introduction}
\label{sec:introduction}
\begin{figure*}
    \centering
    \includegraphics[width=1\linewidth]{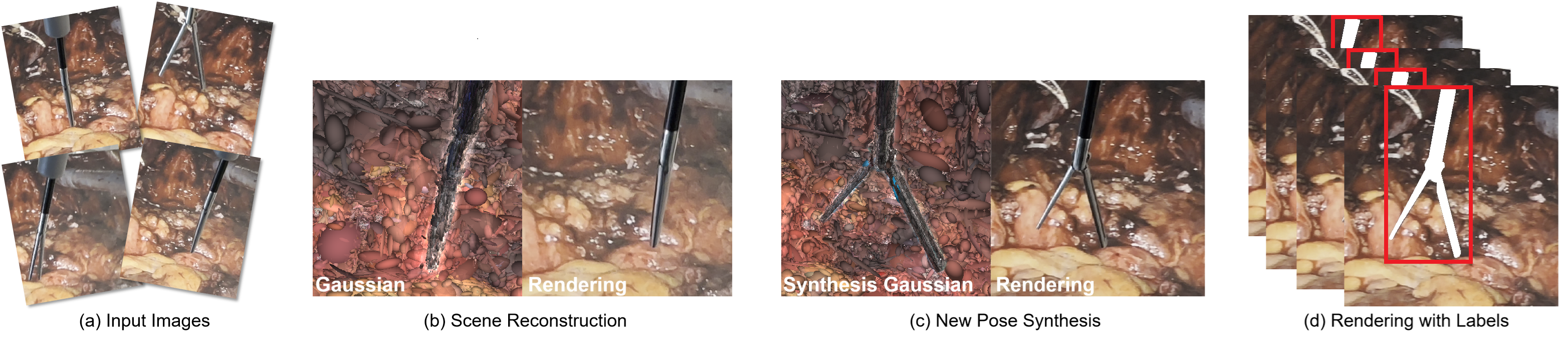}
    \caption{\textbf{Method Overview.} (a) Our method takes images of surgical instruments in various poses as input. (b) Then we reconstruct the 3D Gaussian representation of the surgical scene. (c) After the reconstruction, our method can predict the 3D Gaussian representation of unseen deformations of the surgical instruments. (d) We render them as realistic images, and additionally can obtain the tool annotations automatically. In (d) the red boxes indicate the bounding boxes, and the white areas represent the instrument segmentation masks.}
    \label{fig:methodoverview}
\end{figure*}

\IEEEPARstart{C}{omputer} vision plays a crucial role in advancing intelligent surgical navigation, planning, and automation \mbox{\cite{mascagni2022computer}}. The effectiveness of deep learning-based surgical vision models heavily depends on the availability of high-quality, structured datasets, particularly for tasks such as instrument tracking, segmentation, and pose estimation. However, obtaining large-scale, high-quality labeled datasets remains a significant challenge in the medical imaging field.

\hl{Most deep learning-based methods used in surgical data science are supervised, requiring a substantial amount of annotated image data for training to ensure robustness in complex and highly dynamic surgical scenarios. Although semi-supervised, self-supervised, and few-shot learning approaches exist, they often face challenges in achieving the same level of performance as fully supervised methods \mbox{\cite{ramesh2023dissecting}}. However, the lack of high-quality labeled surgical image datasets has constrained the development of these methods \mbox{\cite{10005544}}. This scarcity arises from several factors: ethical concerns in recording surgical videos complicate the management and sharing of medical data, while issues such as limited field of view, lens distortion, and frequent obstructions from tools, blood, and smoke during procedures lead to poor image quality and incomplete visual information \mbox{\cite{upadhyay2024advances}}. These challenges make it difficult to generate high-quality datasets, as detailed annotations require significant time and expertise and are further impacted by annotator subjectivity.

To address the data scarcity issue, various approaches have been explored. Some studies have leveraged 3D virtual simulators to generate synthetic images \mbox{\cite{lungu2021review}} or employed photo rendering software \mbox{\cite{liu2020photoshopping}} to create realistic textures. However, these images often lack sufficient realism and diversity. Alternatively, SLAM-based methods \mbox{\cite{luegmair2015three}}, \mbox{\cite{mahmoud2018live}} reconstruct static surgical scenes but lack the capability to generate structured labels or diverse datasets for deep learning training. Other studies have attempted data augmentation via image synthesis \mbox{\cite{garcia2021image}}, weak annotations \mbox{\cite{fuentes2019easylabels}}, or generative models using Image-to-Image Translation techniques \mbox{\cite{zhu2017unpaired}}. While these methods improve dataset availability, they struggle to capture the complexity and variability required for real-world surgical scenarios.

Recent advancements in Neural Radiance Fields (NeRFs) \mbox{\cite{mildenhall2021nerf}} and 3D Gaussian Splatting (3DGS) \mbox{\cite{kerbl20233d}} have introduced new opportunities for dataset generation. Both methods can model complex 3D scenes, and recent approaches that integrate time-dependent neural displacement fields \mbox{\cite{liu2024endogaussian}}, \mbox{\cite{wang2022neural}} have become representative works in dynamic surgical scene reconstruction. Compared to NeRF, 3DGS provides explicit scene representation, allowing for efficient rendering and scene-editing capabilities. This makes 3DGS particularly suitable for reconstructing surgical environments. However, most 3DGS-based methods focus on modeling tissue deformation rather than generating structured datasets for deep learning applications. These methods typically rely on continuous temporal input to reconstruct previously observed tissue deformations, limiting their ability to synthesize novel instrument states or automatically generate labeled datasets.

To address these challenges, we propose NeeCo, a novel framework designed for generating high-quality, labeled surgical image datasets. Unlike prior 3DGS-based methods that primarily focus on reconstructing dynamic surgical environments, NeeCo is explicitly designed to generate diverse and structured datasets by leveraging instrument kinematics and Gaussian-based rendering. By learning instrument motion from unordered images, NeeCo enables the synthesis of novel instrument states, significantly expanding dataset diversity while ensuring high annotation accuracy.}

Our contributions can be summarized as follows:

1. \textbf{Novel Dynamic Surgical Instrument Reconstruction Framework:} \hl{We propose an innovative framework that synthesizes novel scenes of kinematically-posable surgical instruments, learned from previously observed images of recorded instrument kinematics (position, rotation, and jaw aperture angle). This framework can reconstruct instruments in dynamic surgical scenes and predict the view under tool movement, including unseen pose and position changes for instruments.}

2. \textbf{Dynamic Adjustment Method for 3DGS Training:} \hl{We introduce a method for dynamically adjusting the 3DGS training process. We do this by adopting different training strategies at various stages. Our approach addresses the challenges posed by poor initialization from inaccurate camera poses when using our recorded surgically relevant scenes of ex-vivo porcine organs as input.} 

3. \textbf{Automatic Generation of Annotations:} The proposed method can automatically generate annotation information alongside the rendered images without human intervention based on the dynamic Gaussian kinematic changing.

4. \textbf{Evaluation with Ground Truth (GT) Images of Anatomy:} We record a new dataset using ex-vivo abdominal organs from a porcine model while tracking our tool state, producing GT images and data that can be directly compared with our generated images. Through comparative experiments on image quality and neural network training using GT images as the benchmark, we ensure the reliability of our conclusions and show a marked increase in synthesis quality and compared with comparable SOTA methods.

\section{Related Work}
This section examines significant advancements in surgical scene reconstruction, focusing on traditional, implicit, and explicit representation methods. Given that our approach emphasizes the reconstruction and representation of surgical scenes that include instruments, we also review current methodologies for surgical instrument synthesis in medical imaging.
\subsection{Surgical Scene Reconstruction}

\subsubsection{Traditional Representation}

Early studies, such as \cite{luegmair2015three}, \cite{mahmoud2018live}

relied on stereo inputs to recover scene depth information through SLAM techniques. These methods generated depth maps via depth estimation and fused the depth maps from multiple viewpoints in 3D space to achieve static scene reconstruction. Subsequent advancements, such as \cite{li2020super} and \cite{long2021dssr}, introduced new stereo depth estimation frameworks by tracking the deformations of key points in the scene, enabling simple 3D deformable reconstruction. These methods heavily rely on sparse deformation fields for tracking deformations and are limited when faced with complex or significant deformations, capturing only relatively small changes. Additionally, the overall quality of the images generated by these methods is often subpar, further limiting their effectiveness in training robust neural networks.
\subsubsection{Implicit Representation}
Implicit representations, such as NeRF \cite{mildenhall2021nerf}, have significantly advanced medical imaging. Unlike traditional methods that rely on spatial geometric information and the tracking of key deformation points for reconstruction, implicit representations use neural radiance fields and deformation fields to capture and represent scene deformation. This combination facilitates the effective reconstruction of dynamic scenes. Recently, EndoNeRF \cite{wang2022neural}

has emerged as a promising solution for dynamic surgical scene reconstruction. It uses tool-guided ray casting, stereo depth-cueing ray marching, and stereo depth-supervised optimization to achieve high-quality results but suffers from lengthy training times. To address this, Forplane \cite{yang2023neural} optimizes training by conceptualizing surgical procedures as 4D volumes, decomposed into static and dynamic fields with orthogonal neural planes, reducing memory usage and accelerating optimization. However, these methods neglect the modeling of surgical instruments and produce non-editable, limited generalization scenes.
\subsubsection{Explicit Representation}
\hl{Explicit scene reconstruction methods, such as 3DGS \mbox{\cite{kerbl20233d}}, overcome the limitations of implicit representation methods, which are difficult to edit. By manipulating the obtained 3D Gaussians, objects can be rotated and translated without sacrificing reconstruction quality. Additionally, these methods enable rapid training and real-time rendering of reconstructed scenes. Similar to EndoNeRF \mbox{\cite{wang2022neural}}, EndoGaussian \mbox{\cite{liu2024endogaussian}}, and EndoGS \mbox{\cite{zhu2024deformable}}, which use 3D Gaussians to represent surgical scenes, recent works have further adapted 3DGS for surgical applications.

Several studies have integrated 3DGS into dynamic endoscopic reconstruction by incorporating temporal deformation fields and depth priors. Deform3DGS \mbox{\cite{yang2024deform3dgs}} models soft tissue deformations at the level of individual Gaussians, while Endo4DGS \mbox{\cite{huang2024endo}} and HFGS \mbox{\cite{zhao2024hfgs}} improve tissue motion reconstruction by leveraging depth priors and frequency-based optimization. While these methods enhance real-time rendering and tissue dynamics modeling, they focus primarily on deformable surgical environments rather than instrument state representation or unseen pose synthesis.

To improve camera pose estimation, Free-SurGS \mbox{\cite{guo2024free}} introduces an SfM-free optimization scheme using optical flow priors. EndoGSLAM \mbox{\cite{wang2024endogslam}} further integrates SLAM-based tracking with 3DGS to enhance real-time visualization in endoscopic surgeries. However, both methods are tailored for static scene reconstruction and do not modify the core 3DGS training process to handle dynamic surgical tools.

Gaussian Pancakes \mbox{\cite{bonilla2024gaussian}} improves static tissue reconstruction by incorporating geometric and depth regularization to enhance Gaussian alignment and texture fidelity. While effective for colonoscopic imaging, it does not address the dynamics of surgical instruments or unseen pose synthesis.

Compared to these approaches, our method focuses on instrument-centric scene representation, enabling novel tool pose synthesis and automatic label generation. By introducing a dynamic training adjustment mechanism, our method enhances 3DGS robustness to pose inaccuracies, facilitating accurate reconstruction of dynamic surgical instruments, which is crucial for dataset augmentation and downstream learning-based models.}

\begin{figure*}[t!]
    \centering
    \includegraphics[width=1\linewidth]{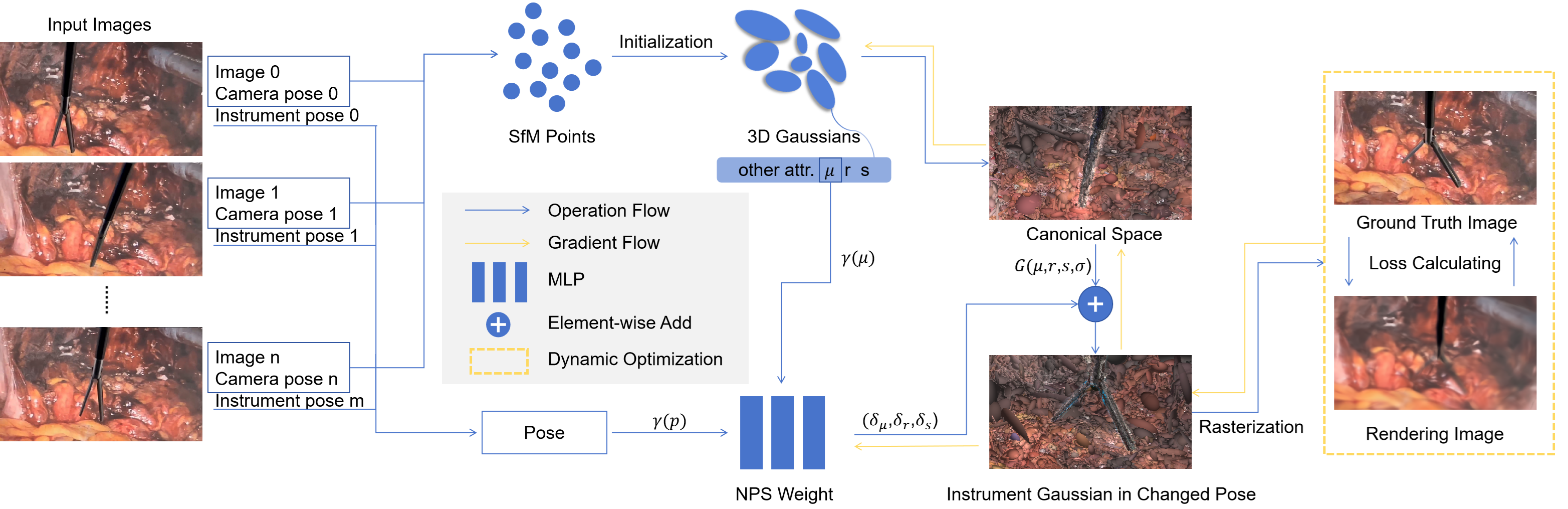}
    \caption{\textbf{Training Process Overview.} Given a set of unordered images of a laparoscopic surgery procedure, our method represents the dynamic scene using 3D Gaussians. It synthesizes the Gaussian representation of the instrument in a new pose from novel viewpoints. The standard 3D Gaussian scene representation is trained in a canonical space. An MLP estimates the attribute changes during scene deformation, transforming the canonical Gaussians to the new deformation. The transformed Gaussians are then rendered using rasterization. The rendered images are compared with ground truth images to evaluate the training. The entire training process is optimized using a dynamic optimization strategy.}
    \label{fig:trainingoverview}
\end{figure*}

\subsection{Instrument Synthesis in Medical Imaging}
Various methods have been developed to generate surgical instrument images. Game engines and surgical simulators \cite{moore2023interactive}, \cite{grossi2023video} provide scalable, noise-free solutions that automatically generate annotations. However, the synthetic images produced by these methods often lack realism, failing to accurately replicate the lighting and textures of real scenes, which can negatively impact the generalizability of models trained on such data. Generative neural networks like GANs and Cycle-GANs \cite{sahu2020endo}, \cite{sahu2021simulation}, \cite{zhang2021surgical} can create synthetic datasets that closely resemble real image distributions, achieving higher image quality. However, these methods can typically only generate pseudo annotations for tasks like segmentation, and they often fall short of providing the high-quality, accurate annotations required for supervised learning. Recent approaches \cite{colleoni2022ssis}, \cite{9976185} have attempted to integrate simulation environments with generative networks, enhancing simulated medical images with real image characteristics to produce high-quality, annotated synthetic datasets. Nevertheless, these methods are primarily designed for static scenes and struggle to modify the pose, orientation, and deformation of instrument end-effectors, which limits their effectiveness in dynamic surgical environments.

\hl{Recent work such as Instrument-Splatting \mbox{\cite{yang2025instrument}} demonstrates controllable, photorealistic rendering of previously unseen tool poses and jaw configurations from monocular surgical videos, enabling high-quality synthetic data generation. However, its reliance on precise, part-level CAD models to bind Gaussian splats to instrument geometry limits accessibility, as such models are often proprietary and unavailable in many surgical domains.}

\section{Methodology}
We propose a method for synthesizing realistic images of surgical instruments exploiting the recent advances in 3D Gaussian Splatting. Our approach allows for dynamic selection of both the user's viewpoint and the instrument's kinematic state (6-DoF pose and jaw aperture angle), allowing for a user to train a 3DGS model on a limited size dataset before synthesizing supplemental images to enhance the performance of neural networks trained on the data. Our model is trained from a monocular video of a moving surgical instrument in a surgical scene; each frame is recorded with kinematic information of the tool's current state. From this, we train a deformable Gaussian model using an MLP to decode user-specified instrument states. Our approach utilizes multiple enhancements to the training method of deformable Gaussian models, namely dynamically adjusting the training rates of the Gaussian's properties, uniform motion rendering, and dynamic compensation of camera poses. We also use our trained Gaussian model to automatically generate segmentation masks and relevant instrument bounding boxes before proving the quality of our image synthesis to enhance the performance of AI-based tool detection and segmentation.
\subsection{Preliminary: 3D Gaussian Splatting}
3DGS utilises a field of explicitly defined Gaussians in 3D space to define and render a 2D scene \cite{kerbl20233d}. We initialize our 3D Gaussians from point clouds generated by COLMAP \cite{schonberger2016structure}, following the specified mathematical expression:
\begin{equation}
G(x) = e^{-\frac{1}{2}(\bm{x}-\bm{\mu})^T \bm{\Sigma}^{-1} (\bm{x}-\bm{\mu})}
\end{equation}
where \( \bm{\mu} \) denotes the mean value of the point cloud \(P(x,y,z)\) and \( \bm{\Sigma} \) is a 3D covariance matrix, expressed as \(\bm{\Sigma = R S S^T R^T}\). Here, \(\bm{R}\) denotes a \(3 \times 3\) rotation matrix, and \(\bm{S}\) is a \(3 \times 3\) diagonal matrix representing the scale. To simplify its representation, the rotation matrix \(\bm{R}\) is converted into a vector \(\bm{r}\).
 These 3D Gaussians are projected into 2D and rendered for each pixel using the following 2D covariance matrix $\Sigma'$:
\begin{equation}
\bm{\Sigma'} = \bm{J W \Sigma W^T J^T},
\end{equation}
where \bm{$J$} is the Jacobian of the affine approximation of the projective transformation, \bm{$W$} is the view matrix transitioning from world to camera coordinates, and \bm{$\Sigma$} denotes the 3D covariance matrix.

The color of the pixel on the image plane, denoted by $\mathbf{C}$, is calculated by \(\alpha\)-blending the contributions of the \(N\) Gaussians, which are sorted from closest to farthest:
\begin{equation}
\mathbf{C} = \sum_{i \in N}\alpha_i c_i\prod_{j=1}^{i-1} (1 - \alpha_j)
\end{equation}
\begin{equation}
\alpha_i = \sigma_i e^{-\frac{1}{2} (\bm{\mu} - u_i)^T \Sigma' (\bm{\mu} - u_i)}
\end{equation}
where $c_i$ represents the color of each Gaussian along the ray, and $u_i$ denotes the $uv$ coordinates of the 3D Gaussians projected onto the 2D image plane.

During initialization, the Gaussian is also assigned an opacity attribute \(\sigma\), thus the 3D Gaussian is defined as: 
\begin{equation}
G(\bm{\mu}, \bm{r}, \bm{s}, \sigma)
\end{equation}

\subsection{Deformable Gaussian Splatting}
When part of the scene undergoes deformations, the Gaussian properties in the deformation regions are varied to represent the new state, expressed as:
\begin{equation}
G'(\bm{\mu}', \bm{r}', \bm{s}', \sigma') =  G(\bm{\mu} + \delta \bm{\mu},  \bm{r} + \delta \bm{r}, \bm{s} + \delta \bm{s}, \sigma + \delta \sigma) 
\end{equation}
To mitigate the impact of dynamic changes in the scene, we train this standard Gaussian model in the canonical space. To decode from our canonical model \(\mathcal{C}\), we use an MLP to model the required parametric change in \( \bm{\mu}, \bm{r}, \) and \( \bm{s} \) for each Gaussian, referred to as the New Pose Synthesis (NPS) weight \(\mathcal{F}\), with depth and hidden layer size of \( D = 12 \) and \( W = 256 \), respectively. The NPS weight takes two inputs: a representation of object movement in the scene, $\mathbf{p}$, and the coordinates of the Gaussian centers in the previous frame, $\bm{\mu}$. The parameter $\bm{\mu}$ primarily represents positional changes (translation), while $\mathbf{p}$, which includes the rotation and jaw aperture change of the laparoscopic instrument, captures the instrument's deformation. Together, these two parameters effectively represent the scene's deformations. We use a positional encoder ($\gamma$) on both $\mathbf{p}$ and $\bm{\mu}$ (adapted from \cite{mildenhall2021nerf}) to enhance training quality, where:
\begin{equation}
\gamma (\mathbf{\bm{\mu}}) = (\sin (2^k \pi \mathbf{\bm{\mu}}), \cos (2^k \pi \mathbf{\bm{\mu}}))_{k=0}^{L-1}
\end{equation}
Resulting in:
\begin{equation}
(\delta \bm{\mu}, \delta \bm{r}, \delta \bm{s}) = \mathcal{F} (\gamma (\bm{\mu}), \gamma (\bm{p}))
\end{equation}
And giving the final representation of the synthesised scene as:
\begin{equation}
G'=\mathcal{C} + \mathcal{F} ( \gamma ( \bm{p} ), \gamma ( \bm{\mu} ) )
\end{equation}

 We do not update the opacity \(\sigma\) during this process, as it primarily affects the rendering process by determining the final rendered color. Since instruments typically keep to a constant colour space regardless of pose, we do not estimate the opacity. Nevertheless, \( \sigma \) of the Gaussian in the canonical space is still optimized during training.\\

It is worth noting that methods like 4DGS \cite{wu20244d} and D-3DGS \cite{yang2024deformable} also use MLPs to model dynamic scenes. However, in these methods, the primary aim of the MLP is to decouple continuous dynamic scenes into multiple static scenes. The MLPs in these methods use a time parameter \(t\) to reconstruct learned scenes within the timeframe of the parent video. In essence, they do not treat the dynamic scene as a whole; instead, the MLP learns the Gaussian representation of each decoupled static scene individually, requiring the input data to be temporally continuous and within range of the training data. In contrast, the MLP in our approach is designed to predict changes in Gaussian attributes based on object movement, training and learning from the dynamic scene as a whole. Consequently, our method can handle unordered image inputs and generate scenes from object movement outside of the observed values.

\subsection{Deformable Gaussian Training}
The input is unordered images that capture a dynamic surgical scene with instrument deformation. Initially, we use SfM to calibrate the camera pose and generate a sparse point cloud representing the scene, along with the object movement parameter \( \bm{p} \) for each frame. The sparse point cloud is then initialized into Gaussians and transferred into the canonical space for training. As the scene transitions from between frames, the Gaussian attributes \( \bm{\mu} \) and \( \bm{p} \) of the current frame are encoded and fed into the NPS weight, which attributes necessary to transform the Gaussians from the current frame to the next. 

As training progresses, the NPS weight gradually learns to induce changes in the scene's Gaussian representation. As this is built from changes in \(\bm{\mu}\) and \(\bm{p}\) between two frames, we can generate new frames by inputting arbitrary \(\bm{\mu}\) and \(\bm{p}\) values. We further validate this ability to generate high-quality Gaussians representing unseen scene transformations in the experimental section \ref{exp}. Following the 3DGS \cite{kerbl20233d}, we render the transformed scene's Gaussians into an image. This rendered image is then compared with the ground truth image of the transformed scene to calculate the loss function \(\mathcal{L}\), a combination of \(\mathcal{L}_1\) loss and a D-SSIM term:
\begin{equation}
\mathcal{L} = (1 - \lambda) \mathcal{L}_1 + \lambda \mathcal{L}_{\text{D-SSIM}}
\end{equation}
It is important to note that the rendering process heavily depends on accurate camera poses. SSIM (Structural Similarity Index Measure) is sensitive to spatial relationships within the image, and viewpoint discrepancies can lead to significant visual differences and cause SSIM to drop sharply.

Therefore, we assign a higher weight (\(1-\lambda = 0.9\)) to the \(\mathcal{L}_1\) term, reducing the weight of the \(\mathcal{L}_{\text{D-SSIM}}\) term and minimizing the impact of camera pose errors on the training outcome. After calculating the loss, we update the attributes of the Gaussians in the canonical space and all NPS weight parameters. Fig.~\ref{fig:trainingoverview} summarizes our training process.
\subsection{Dynamic Training Adjustment}

\begin{figure}[t]
    \centering
    \includegraphics[width=1\linewidth]{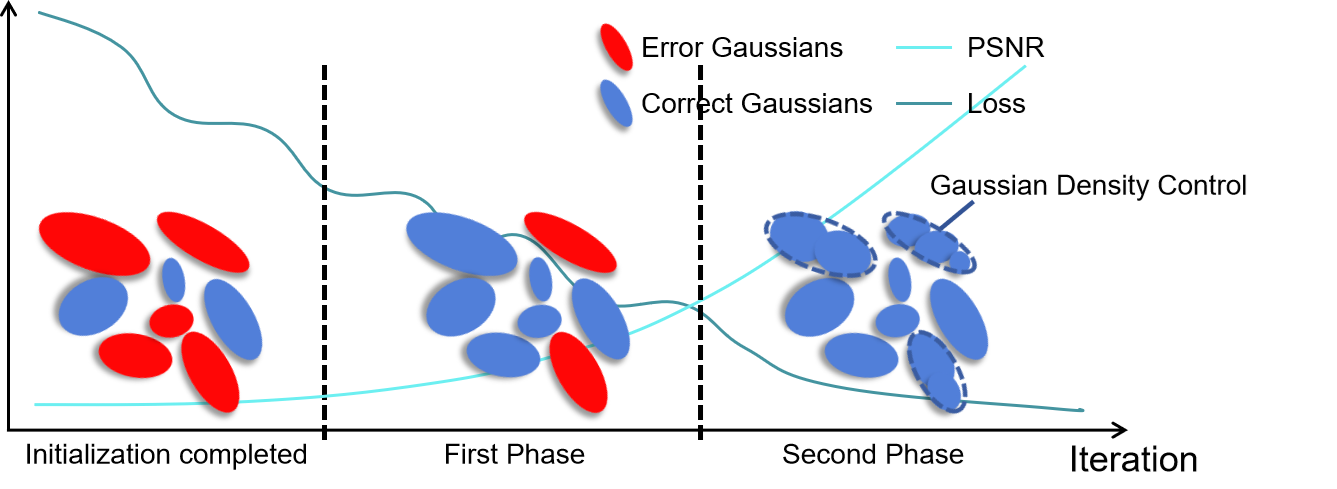}
    \caption{\textbf{Dynamic Density Control.} In the first phase of training, the Gaussian distributions are newly initialized and contain numerous error points. During this phase, the loss function shows fluctuations, and the PSNR values are low but steadily increasing. In the second phase, the density control is gradually relaxed throughout the iterations, allowing for the splitting and cloning of Gaussian points to further enhance the training quality.}
    \label{fig:density_control}
\end{figure}
\subsubsection{Dynamic Density Control Adjustment}\label{DDCA}

Density control is essential to the 3DGS pipeline; it enhances scene detail by cloning small Gaussians in sparse areas and splitting large ones in dense areas. It is predefined before training by three parameters: the densification interval \(P_{di}\), the opacity reset interval \(P_{oi}\), and a positional gradient threshold \(\tau_{pos}\). It is, however, particularly sensitive to initialization errors. Incorrect camera poses introduce numerous erroneous points into the initialized Gaussian models, and density control can exacerbate these errors by misplacing or redundantly multiplying error-prone Gaussians, eventually causing training failure. While 3DGS can correct errors in static scenes, its robustness is compromised with dynamic datasets, as premature density control amplifies errors beyond its corrective capabilities. These fixed parameters do not accommodate the rapid dynamics of real-world scenes. Removing or delaying density control can mitigate these issues but may compromise the overall quality of the Gaussian representation, as density control is crucial for enhancing fidelity and detail.

We propose an adaptive density control strategy, illustrated in Fig.~\ref{fig:density_control}. We partition the training into two phases, guided by reductions in the loss function and improvements in the Peak Signal-to-Noise Ratio (PSNR) of rendered images, and dynamically adjust the density control during the training process. Existing works \cite{zhang2024pixel, bulo2024revising} also explored the density control module in Gaussian Splatting, focusing on enhancing image detail. However, they do not address the negative impact of erroneous initialization on the training process.

In the first phase, Gaussians are newly initialized and often contain numerous errors. During this initial phase, we restrict density control to prioritize correcting these erroneous points to their accurate positions. We extend densification and opacity reset intervals and increase the gradient threshold, setting \( P_{di} = 500 \), \( P_{oi} = 10,000 \), and \(\tau_{pos} = 0.0004\). Once the erroneous points are largely corrected in the first phase, the second phase commences as PSNR values exceed 20 and the loss function consistently declines. Here, we reintroduce density control, enhancing geometric detail and refines overlapping areas, significantly improving model accuracy and robustness. Parameters for this phase are set to \( P_{di} = 200 \), \( P_{oi} = 3000 \), and \(\tau_{pos} = 0.0002\).

\subsubsection{Dynamic Spherical Harmonics Function Update}
\hl{Similar to dynamic density control, we adopt an adaptive strategy for updating Spherical Harmonics (SH) coefficients, aiming to balance training stability and rendering quality. SH functions capture illumination and fine-grained scene details, where higher-order SH terms enhance expressiveness but also introduce increased computational complexity and sensitivity to initialization errors. In standard 3DGS training, SH updates typically follow a fixed schedule (e.g., increasing the SH degree every 1000 iterations). However, this approach does not account for the impact of accumulated errors in dynamic scenes, which can lead to instability during training.
To address this, we propose a stage-based dynamic SH update strategy. During the early training phase, we restrict SH order growth, prioritizing the correction of erroneous Gaussian initializations. This allows the model to focus on learning fundamental geometric structures and global illumination patterns, reducing the risk of error propagation. Once the model stabilizes in later training stages, we progressively increase the SH order, enabling the capture of more intricate lighting variations and texture details.
This staged update strategy ensures that SH refinement does not interfere with early-stage optimization while fully leveraging higher-order SH capabilities for high-precision scene reconstruction. Compared to fixed SH update schedules, our method effectively mitigates initialization errors, improves model robustness in dynamic surgical instrument reconstructions, and leads to higher-quality rendering results.}

\begin{figure}
    \centering
    \includegraphics[width=0.9\linewidth]{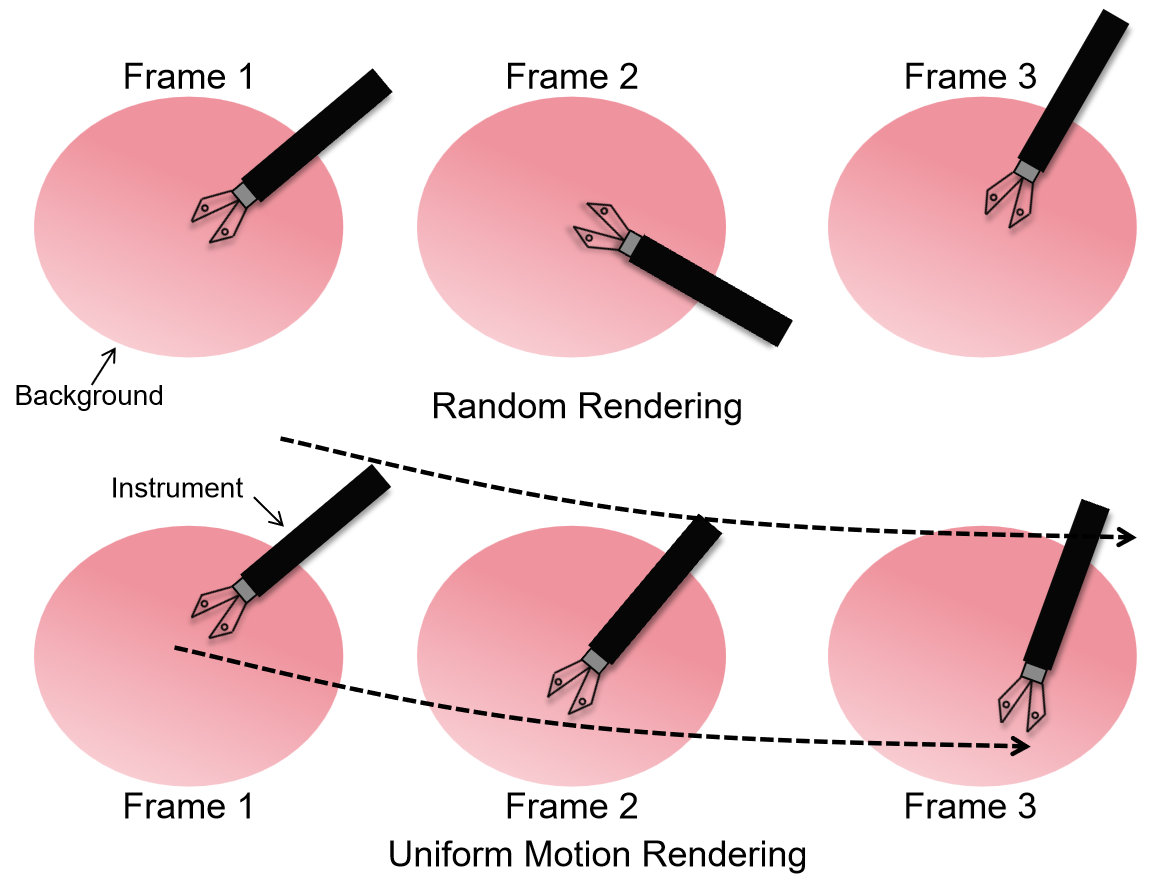}
    \caption{\textbf{Uniform Motion Rendering:} The first row shows random rendering, where the instrument undergoes significant changes between frames. The second row shows uniform motion rendering, where the instrument's movement between frames is more consistent.}
    \label{fig:unimotion}
\end{figure}

\subsubsection{Uniform Motion Rendering}
In the original 3DGS training process, input images are randomly selected during training. However, since our input images include different instrument poses, randomly selected images present significant variations in the instrument's appearance between two consecutive images. An instrument part visible in one frame might become occluded in the next. These substantial variations pose challenges for the NPS Weight to accurately predict Gaussian changes in the early stages of training, causing a larger computed loss and slowing down the convergence of the training process.

To address this issue, we introduce uniform motion simulation as shown in Fig. \ref{fig:unimotion}. In the first training phase, we sort the images according to the pose positions of the instrument, simulating a uniform motion with slow changes. These smaller object motions ($\bm{p}$) are easier to learn, enabling the training process to converge more quickly. Once the training progresses to the second phase, we revert to the random rendering strategy, allowing the model to adapt to larger variations in the instrument's poses. Gradually introducing larger variations helps the NPS learn and adapt to complex changes in the instrument, thereby enhancing the overall performance and robustness of the model.

\subsubsection{Dynamic Camera Pose Compensation}
Training with imprecise poses may negatively impact the training process. Inaccurate camera poses in real-world datasets can cause spatial jitter between frames in the test or training set, resulting in significant deviations between the rendered test images and the ground truth \cite{park2021hypernerf}. We propose a compensation mechanism for the rendering jitter caused by inaccurate camera poses. Previous works \cite{yang2024deformable,seiskari2024gaussian} either directly compensate for camera poses using estimates from visual-inertial odometry (VIO) or are specifically designed for tasks involving temporal interpolation. In contrast, our compensation method can adapt to errors introduced by camera movements in all directions and is applicable in stages, aligning seamlessly with the dynamic training process outlined in this work. This compensation primarily occurs in the first phase of training: 
\begin{equation}
    (\delta \bm{\mu}, \delta \bm{r}, \delta \bm{s}) = \mathcal{F}(\gamma(\mathbf{\bm{\mu}}), \gamma(\bm{p}) + \Delta)
\end{equation}
\begin{equation}
    \Delta = (\mathcal{N}(0,\sigma^2) \cdot \beta \cdot t_{phase}
\end{equation}
where $\Delta$ represents the compensation\hl{ term, modeled as a zero-mean Gaussian distribution $\mathcal{N}(0, \sigma^2)$ to simulate random camera pose errors. This allows for the modeling of both positive and negative deviations, ensuring a more realistic approximation of real-world camera motion noise. The variance $\sigma^2$ is empirically determined to balance training stability, while $\beta$ serves as a scaling factor to control the magnitude of the perturbations. The boolean variable $t_{phase}$ determines the training phase in which the compensation is applied, with a value of 1 during the first phase and 0 during the second phase.}

\subsection{Automatic Annotation Generation}

We generate dataset annotations in two steps. First, we generate a segmentation mask, and second, we use this to define a bounding box for detection. When instrument movement occurs, Gaussians (\(G\)) in the scene may be represented as \(G'(\bm{\mu} + \delta\bm{\mu}, \bm{r}+\delta \bm{r}, \bm{s}+\delta \bm{s}, \sigma)\). When rendered with a new pose, the NPS weights (\(\delta\bm{\mu}, \delta \bm{r}, \delta \bm{s}\)) for the instrument (the moving part of the scene), are much greater than those observed in the background (mainly static). To identify significant changes, we experimentally establish a variation threshold \(\mathcal{H}_{\delta \bm{\mu}},\mathcal{H}_{\delta r}, \mathcal{H}_{\delta s} \). If the increment in Gaussian attributes exceeds this threshold, we consider it indicative of substantial deformation, labelling these Gaussians as constituents of the instrument.

By rendering only the Gaussians corresponding to the surgical instrument, we can effectively generate a segmented image of the tool: the background is devoid of Gaussians and will consequently be rendered black, while the instrument is colored as normal (with some artifacts around the edges as a result of low opacity regions of the perimetric Gaussians. We set a threshold of the pixel's RGB magnitude on this segmented rendered image, generating a binary segmentation mask. With this mask, we can automatically generate minimum bounding boxes using the mask's outer contour and bounding boxes using the extremity pixel coordinates.

\section{Experiment}
\label{exp}
\subsection{Data Recording and Experimental Setup}
\subsubsection{Dataset Collection}
\begin{figure}[t]
    \centering
    \includegraphics[width=1\linewidth]{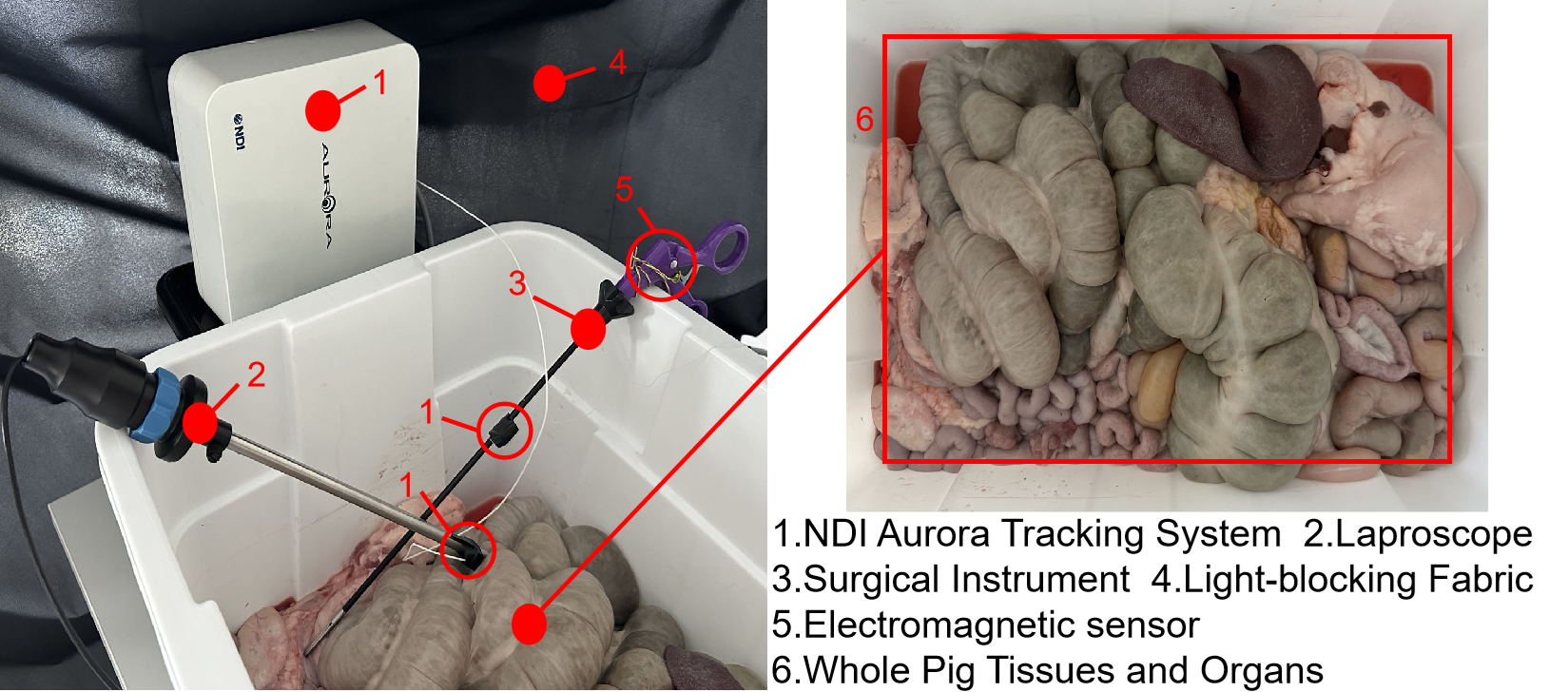}
    \caption{Data recording platform.}
    \label{fig:expsetup}
\end{figure}

\hl{Publicly available surgical datasets, such as SCARED \mbox{\cite{allan2021stereo}} and EndoNERF \mbox{\cite{wang2022neural}}, are widely used for endoscopic scene reconstruction. However, these datasets are not suitable for our method due to critical limitations. The SCARED dataset primarily focuses on static tissue reconstruction and lacks dynamic surgical instruments, which are essential for modeling instrument motion and deformation. EndoNERF, on the other hand, does not provide kinematic data of dynamic surgical instruments, making it unsuitable for our approach, which relies on 7-DoF motion parameters for accurate instrument representation and unseen pose synthesis. Given these limitations, existing datasets do not provide the necessary information required for our study, prompting us to construct a dedicated experimental platform for data collection.}
Our experimental setup consists of a laparoscope (Endoskill, MedEasy, China), laparoscopic instruments (MedEasy, China), an electromagnetic (EM) tracking system (Aurora, NDI, Canada), allowing 6-DoF motion tracking of the instrument and laparoscope. Additionally, we modified the instrument handle with a Hall-effect sensor to measure the jaw opening angle. Fig. \ref{fig:expsetup} shows the data recording platform.

We used ex-vivo pig tissue and organs during data collection to create a more realistic surgical environment. Organs were harvested from pigs reared and slaughtered for the food chain. We collected data from various tissues and organs, including the liver, stomach, and colon. These tissues and organs exhibit different visual characteristics such as color, texture, and shape, effectively simulating different surgical environments within the human abdomen. During recording, we used the EM tracker and jaw angle sensor to capture the 7-DoF data of the surgical tools and the camera's 6-DoF pose. We exclusively utilized lighting sources from the laparoscope to achieve a more realistic simulation of real-world surgical scenarios.

Our Gaussian model training dataset consists of three videos, each of a subject organ from our ex-vivo experimental setup (colon, liver, stomach), containing 576,000 frames with associated instrument states. To avoid redundancy, we sampled 14,000 frames which were used as a training and validation set for training the segmentation and detection Neural Networks. We collected an additional dataset as a test set. This dataset contains 300 images featuring various backgrounds, instrument poses, and deformations not seen in our training data.

\subsubsection{Implementation Details}
We implement our Gaussian model training pipeline using Pytorch. All Gaussian models were trained on an NVIDIA RTX Ada 4500 graphics card and Intel Xeon W5-2455X CPU. Aside from those set by our Dynamic Training Adjustment, all other training parameters are set based on the initial 3DGS work.\\
We train our Neural Network evaluators on an NVIDIA RTX 4050(6G) graphics card and Intel i7-12650H CPU. All models were trained using their default parameters for 300 epochs.

\subsubsection{Evaluation Metrics}
For our image quality comparison, we render full images to compare against the GT. Therefore, we utilize three comparative evaluation metrics that compute similarity over the full image: SSIM, PSNR, and LPIPS, typically used within synthetic image generation \cite{kerbl20233d,park2021nerfies,yang2024deformable}. For our Neural Network evaluators, we use Precision and Recall for object detection and IoU and Dice for segmentation. To mitigate the effect of randomness on model performance evaluation, we conduct multi-fold experiments for each model.

\subsection{Image Quality Experiment}

\begin{figure*}[htb]
    \centering
    \includegraphics[width=1\linewidth]{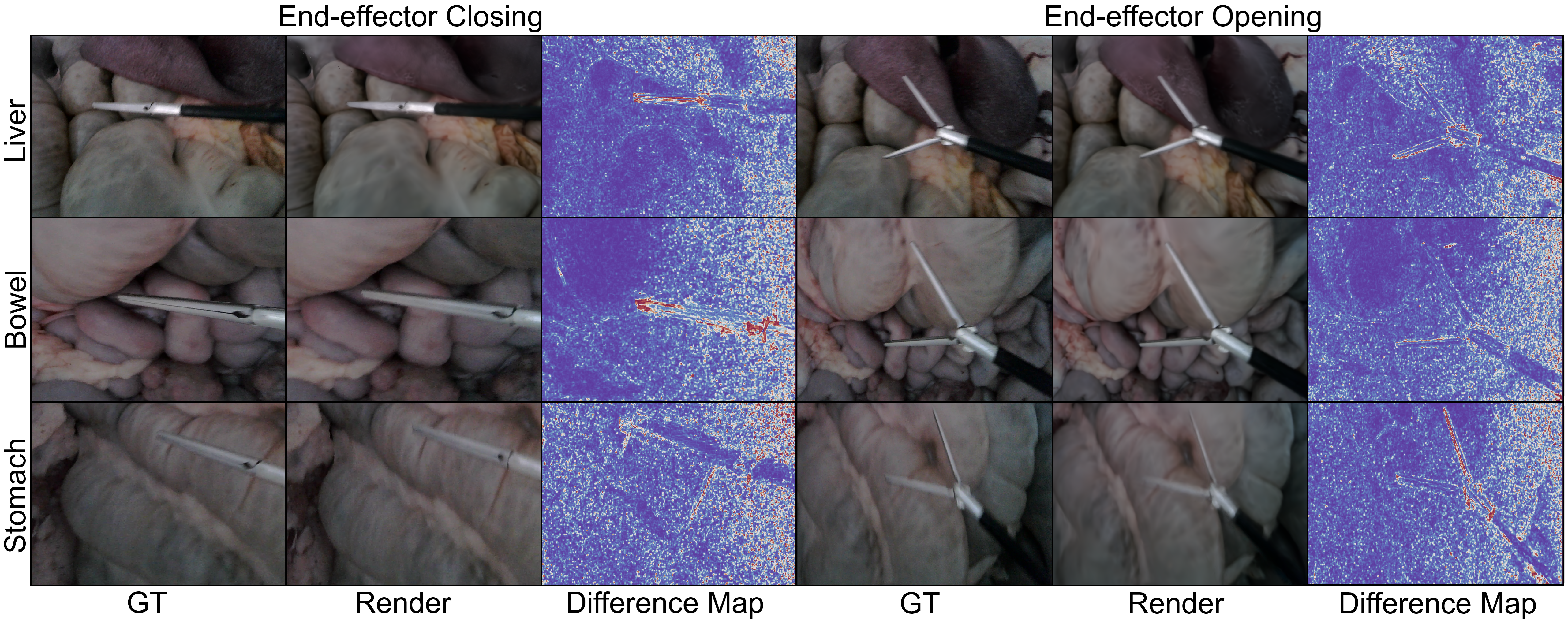}
    \caption{Reconstruction result with difference maps. From left to right are the GT image, the rendered image, and the difference map. The difference map is created by overlaying the two images, with warm colors indicating areas of difference.}
    \label{fig:diffbg}
\end{figure*}

\begin{table*}[htb]
\centering
\caption{Quantitative result of the comparison experiment}
\label{compare_table}
\begin{tabular}{ccccccccccccc}
\\ \toprule
 & \multicolumn{3}{c}{Liver} & \multicolumn{3}{c}{Stomach} & \multicolumn{3}{c}{Bowel} \\
Methods & PSNR $\uparrow$ & SSIM $\uparrow$ & LPIPS $\downarrow$ & PSNR $\uparrow$ & SSIM $\uparrow$ & LPIPS $\downarrow$ & PSNR $\uparrow$ & SSIM $\uparrow$ & LPIPS $\downarrow$ \\
\midrule
3DGS\cite{kerbl20233d} & 18.01 & 0.708 & 0.643 & 17.21 & 0.701 & 0.553 & 18.33 & 0.722 & 0.631 \\
NeRFies\cite{park2021nerfies} & 23.12 & 0.772 & 0.493 & 22.17 & 0.763 & 0.441 & 21.71 & 0.714 & 0.512 \\
4DGS\cite{wu20244d} & 25.23 & 0.847 & 0.411 & 25.41 & 0.837 & 0.391 & 23.41 & 0.786 & 0.428 \\
D-3DGS\cite{yang2024deformable} & 24.01 & 0.811 & 0.462 & 23.32 & 0.803 & 0.422 & 24.68 & 0.835 & 0.431 \\
Unseen Deformation & 27.52 & 0.881 & 0.353 & 27.01 & 0.855 & 0.301 & 27.81 & 0.868 & 0.337 \\
\textbf{NeeCo} & \textbf{28.88} & \textbf{0.902} & \textbf{0.273} & \textbf{29.81} & \textbf{0.893} & \textbf{0.274} & \textbf{29.87} & \textbf{0.913} & \textbf{0.281} \\
\bottomrule
\end{tabular}
\end{table*}
Our image quality experiments evaluated the images rendered by the proposed method, comparing them with the GT (the last three images in Fig. \ref{fig:diffbg}). We replicate our GT images by inputting the GT 7-DoF data, allowing the model to render the identical scene as observed in the GT images (column 1 in Fig.\ref{fig:diffbg}). We overlaid the rendered images with the GT images to generate PSNR difference maps, visualizing the discrepancies between them. As shown in Fig.\ref{fig:diffbg}, the red areas, indicating discrepancies, are primarily located in the detailed regions of the surgical tool jaw and the image edges. 

\begin{figure*}[htb]
    \centering
    \includegraphics[width=1\linewidth]{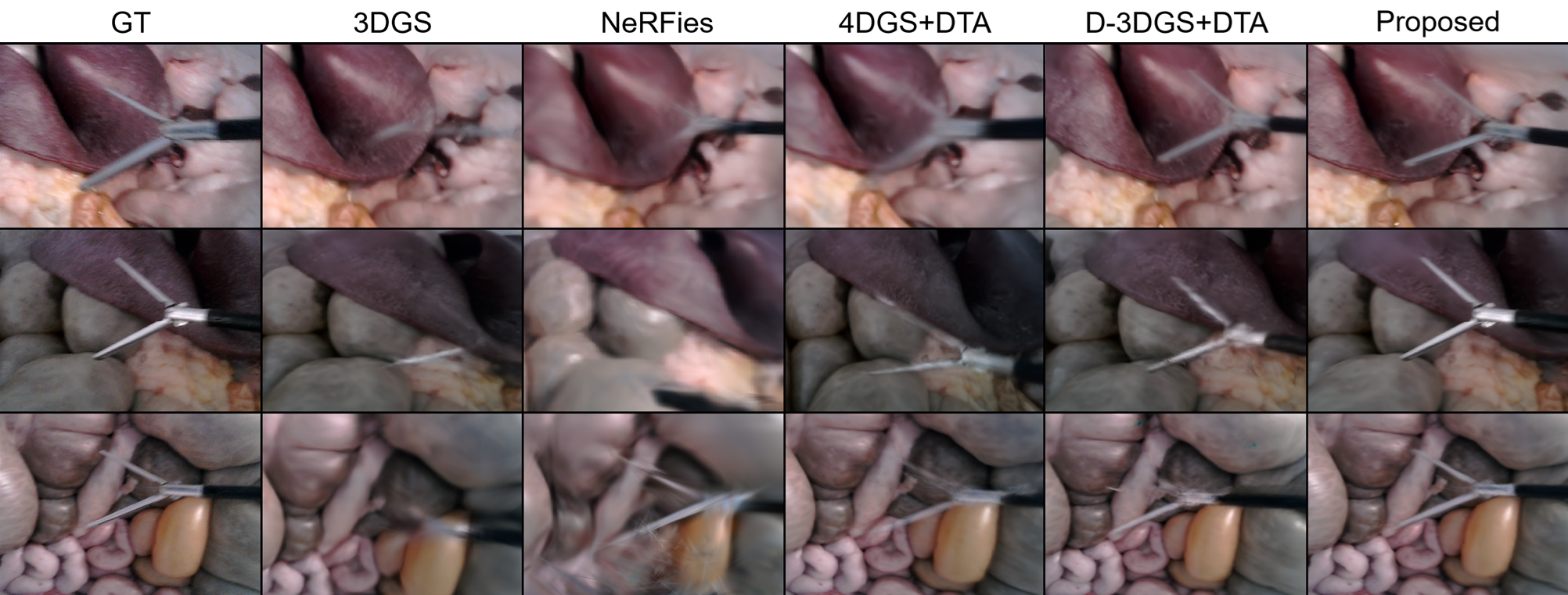}
    \caption{Comparison of dynamic scene reconstruction using various methods. Each row represents the same training dataset, with the proposed method consistently showing more detailed reconstruction results, especially in the main body and jaw parts of the surgical instruments. \hl{The notation ‘+DTA’ indicates that the method has been integrated with the proposed dynamic training adjustment (DTA) strategy.}}
    \label{fig:compare_exp}
\end{figure*}

We selected several state-of-the-art (SOTA) methods for comparison, including 3DGS\cite{kerbl20233d}, NeRFies\cite{park2021nerfies}, 4DGS\cite{wu20244d}, and D-3DGS\cite{yang2024deformable}. \hl{Notably, the original 4DGS \mbox{\cite{wu20244d}} and D-3DGS \mbox{\cite{yang2024deformable}} models could not operate successfully on our dataset due to inaccuracies in camera pose estimation from SfM initialization. To ensure a fair comparison, we integrated our dynamic training adjustment (DTA) \mbox{\ref{DDCA}} strategy into these methods, which enabled their successful application to real-world data and allowed them to converge on our dataset .} The rendering results of all the comparison methods are visualized in Fig. \ref{fig:compare_exp}. Table \ref{compare_table} summarizes the quantitative results on different datasets.

As shown in Table \ref{compare_table}, the rendered images from our proposed method outperform the SOTA methods across various evaluation metrics on different background datasets. Moreover, our method achieves satisfactory results even for unseen deformations. From Fig. \ref{fig:compare_exp}, it can be observed that, although the proposed method exhibits some blurring in certain background regions, it excels in rendering surgical instruments, with both the main body and jaw rendered clearly and closely resembling the GT images.

Among the comparison methods, 3DGS \cite{kerbl20233d} struggles to accurately render static surgical tools in dynamic scenes, while NeRFies \cite{park2021nerfies} can only partially render surgical instruments, producing blurry and unrecognizable results due to it's reliance on highly accurate camera poses, which are difficult to obtain in real-world settings, leads to poor performance with inaccurate poses. 4DGS \cite{wu20244d} and D-3DGS \cite{yang2024deformable}\hl{, when augmented with our proposed dynamic training adjustment strategy, were able to process real-world datasets. However, despite their improved performance, their rendering quality remains inferior to our approach, particularly in capturing unseen instrument deformations and maintaining fine structural details.}

\begin{table}[t]
    \centering
    \caption{\hl{Comparison of annotation accuracy across different segmentation methods}}
    \label{tab:annotation_results}
    \begin{tabular}{lcc}
        \toprule
        \textbf{Method} & \textbf{IoU (\%)} & \textbf{Challenge IoU (\%)} \\
        \midrule
        SAM2\cite{ravi2024sam} & 70.24 & 68.38 \\
        S3NET\cite{baby2023forks} & 79.12 & 77.93 \\
        SurgicalSAM\cite{yue2024surgicalsam} & 83.07 & 83.07 \\
        \textbf{NeeCo} & \textbf{88.21} & \textbf{88.21} \\
        \bottomrule
    \end{tabular}
\end{table}

\subsection{Annotation Evaluation Experiment}\label{sec:annotation_exp}

\hl{To evaluate the accuracy of the automatically generated annotations, we conducted a comparative experiment against existing 2D detection methods. Specifically, we selected two SAM-based \mbox{\cite{kirillov2023segment}} models, SAM2 \mbox{\cite{ravi2024sam}} and SurgicalSAM \mbox{\cite{yue2024surgicalsam}}, along with a specialist model, S3NET \mbox{\cite{baby2023forks}}, which is specifically designed for medical instrument segmentation. Among them, SAM2 is a general-purpose segmentation model trained on a large-scale natural image dataset without fine-tuning for medical applications, while SurgicalSAM and S3NET are designed specifically for surgical instrument segmentation.

The experiment was conducted on a subset of 1000 images, with manual annotations serving as the ground truth. For evaluation, we followed the standard metrics used by these methods, computing IoU and Challenge IoU across multiple datasets. The results are presented in Table \mbox{\ref{tab:annotation_results}}.

As shown in the results, NeeCo achieves the highest segmentation accuracy. This is because NeeCo identifies instrument-related Gaussian ellipsoids directly in the 3D Gaussian space and renders them into 2D labels, avoiding the cumulative errors commonly introduced by 2D detection methods. Consequently, NeeCo achieves the highest IoU and Challenge IoU. In contrast, SurgicalSAM and S3NET operate directly on 2D images. Although both models are specifically optimized for instrument segmentation and achieve relatively good results, NeeCo's automatic annotation strategy, based on 3D Gaussian representations, provides a notable advantage, resulting in superior accuracy. Additionally, SAM2, as a general-purpose segmentation model, lacks fine-tuning on medical datasets, leading to relatively inferior performance.

These findings demonstrate that NeeCo's automatic annotation generation method provides highly accurate labels, surpassing state-of-the-art medical segmentation models. Furthermore, as NeeCo generates these labels automatically within the 3D Gaussian representation, it eliminates the need for manual intervention required by 2D segmentation approaches, making it a more practical and efficient solution for surgical instrument annotation.}
\subsection{Neural Network Training Experiment}
\label{nnexp}
In our neural network training experiments, we validate the effectiveness of the proposed method using rendered images for training neural networks. We focus on two common downstream tasks in RAMIS: object detection and semantic segmentation. We employ well-established and widely-used models in the respective domains: YOLOv5 \cite{redmon2016you} for object detection and U-Net \cite{ronneberger2015u} for semantic segmentation. Additionally, to align with domain-specific benchmarks in medical imaging, we utilize SOTA neural networks specifically designed for surgical tool detection and segmentation. Specifically, we incorporate DBH-YOLO \cite{pan2024dbh} for object detection and RSVIS \cite{wang2024video} for semantic segmentation, leveraging our automatically generated labels.
\begin{table*}[htbp]
  \centering
  \caption{Data source comparison across different model types (Mean and Standard Deviation}
    \begin{tabular}{cccccccc}
    \\ \toprule
      Model & Metrics & REAL & AUGMENT\cite{mikolajczyk2018data} & D-3DGS\cite{yang2024deformable} & 4DGS\cite{wu20244d} & \textbf{NeeCo} & \textbf{NeeCo + Real Images} \\
  
    \midrule 

    \multirow{2}{*}{YOLOv5\cite{redmon2016you}} & Precision $\uparrow$ & 0.703 $\pm$ 0.007 & 0.731 $\pm$ 0.002 & 0.422 $\pm$ 0.009 & 0.476 $\pm$ 0.007 & \textbf{0.694} $\pm$ \textbf{0.013} & \textbf{0.776} $\pm$ \textbf{0.011} \\
     
    & Recall $\uparrow$ & 0.831 $\pm$ 0.003 & 0.852 $\pm$ 0.008 & 0.535 $\pm$ 0.011 & 0.557 $\pm$ 0.006 & \textbf{0.826} $\pm$ \textbf{0.011} & \textbf{0.901} $\pm$ \textbf{0.015} \\

    \midrule
    
    \multirow{2}{*}{DBH-YOLO\cite{pan2024dbh}} & Precision $\uparrow$ & 0.712 $\pm$ 0.007 & 0.731 $\pm$ 0.004 & 0.501 $\pm$ 0.008 & 0.522 $\pm$ 0.006 & \textbf{0.704} $\pm$ \textbf{0.011} & \textbf{0.826 $\pm$ 0.012}\\
    
     & Recall $\uparrow$ & 0.721 $\pm$ 0.004 & 0.744 $\pm$ 0.007 & 0.501 $\pm$ 0.009 & 0.542 $\pm$ 0.008 & \textbf{0.707} $\pm$ \textbf{0.008} & \textbf{0.829} $\pm$ \textbf{0.014} \\

     \midrule
        
    \multirow{2}{*}{U-Net\cite{ronneberger2015u}} & IoU $\uparrow$ & 0.617 $\pm$ 0.008 & 0.622 $\pm$ 0.003 & 0.421 $\pm$ 0.015 & 0.447 $\pm$ 0.009 & \textbf{0.601} $\pm$ \textbf{0.011} & \textbf{0.683} $\pm$ \textbf{0.012} \\
     
    & Dice $\uparrow$ & 0.763 $\pm$ 0.004 & 0.767 $\pm$ 0.009 & 0.592 $\pm$ 0.012 & 0.617 $\pm$ 0.008 & \textbf{0.751} $\pm$ \textbf{0.009} & \textbf{0.812} $\pm$ \textbf{0.013} \\
    
    \midrule
    
     \multirow{2}{*}{RSVIS\cite{wang2024video}} & IoU $\uparrow$ & 0.711 $\pm$ 0.002 & 0.723 $\pm$ 0.003 & 0.581 $\pm$ 0.011 & 0.533 $\pm$ 0.005 & \textbf{0.703} $\pm$ \textbf{0.008} & \textbf{0.817} $\pm$ \textbf{0.009} \\
     
    & Dice $\uparrow$ & 0.831 $\pm$ 0.005 & 0.839 $\pm$ 0.003 & 0.735 $\pm$ 0.007 & 0.695 $\pm$ 0.009 & \textbf{0.826} $\pm$ \textbf{0.004} & \textbf{0.899} $\pm$ \textbf{0.010} \\

    \bottomrule
    \end{tabular}
  \label{compare_nn}
\end{table*}

We train six versions of each model from different source datasets: 1. real images with GT pose,  2. real image augmented by standard data augmentation~\cite{mikolajczyk2018data}. 3 \& 4. synthetic rendering by D-3DGS\cite{yang2024deformable} \& 4DGS~\cite{wu20244d}, respectively, 5. synthetic NeeCo rendering of the GT pose, 6. a mix of NeeCo rendered GT pose and new unseen poses. Each dataset contains 1780 images.

Table \ref{compare_nn} summarizes the performance of different models. Overall, with both DBH-YOLO and \hl{RSVIS}, the GT and models trained on rendered images differ by less than 1.5\% in performance metrics, while the mixed synthetic model shows nearly a 15\% improvement over the GT model. For DBH-YOLO, the GT (Real) and the model trained on the rendered images by proposed method (NeeCo) show similar performance, indicating that networks trained on Render images can match those trained on GT images. Models trained using standard augmentation methods (augment) show performance improvements by rotating or translating the GT images. This helps them handle certain new deformations in the test dataset but are limited in introducing novel views and unseen deformations, leading to only marginal improvements. Although D-3DGS and 4DGS can synthesize new viewpoints, their performance lags due to lower image quality. The mixed synthetic model (NeeCo + real images) outperforms both, as it benefits from unseen deformations and varied camera viewpoints, enhancing training diversity and overall performance, Additionally, the standard augmentation methods can be applied to this model to further boost performance. The conclusions for U-Net mirror those drawn from DBH-YOLO. Similar conclusions can be drawn for YOLOv5 and U-Net. 

\begin{table}[t]
    \centering
    \caption{\hl{Comparison of rendering speed and number of Gaussians}}
    \label{tab:rendering_performance}
    \begin{tabular}{lcccccc}
        \toprule
        \multirow{2}{*}{Method} & \multicolumn{2}{c}{Liver} & \multicolumn{2}{c}{Bowel} & \multicolumn{2}{c}{Stomach} \\
        \cmidrule(lr){2-3} \cmidrule(lr){4-5} \cmidrule(lr){6-7} 
        & FPS & Num(K) & FPS & Num(K) & FPS & Num(K) \\
        \midrule
        D-3DGS\cite{yang2024deformable} & 44 & 140 & 38 & 170 & 33 & 210 \\
        4DGS\cite{wu20244d}   & 38 & 170 & 35 & 200 & 31 & 230 \\
        \textbf{NeeCo}  & \textbf{29} & \textbf{280} & \textbf{24} & \textbf{340} & \textbf{20} & \textbf{380} \\
        \bottomrule
    \end{tabular}
\end{table}

\subsection{Rendering Performance Analysis}

\hl{Table \mbox{\ref{tab:rendering_performance}} compares the rendering FPS and Gaussian point of NeeCo against D-3DGS \mbox{\cite{yang2024deformable}} and 4DGS \mbox{\cite{wu20244d}} across different surgical scenes. While NeeCo exhibits lower FPS (20–29) compared to D-3DGS (33–44) and 4DGS (31–38), this is a direct result of our dynamic density control, which increases the number of Gaussians to achieve higher scene fidelity. For example, in the Liver scene, NeeCo generates 280k Gaussians, nearly double that of D-3DGS (140k), leading to a lower FPS but richer detail representation.

Unlike methods optimized for real-time rendering, NeeCo prioritizes dataset quality over speed, ensuring high-fidelity supervision signals for deep learning models. The increased Gaussian density enhances the representation of surgical scenario, making it particularly suited for dataset synthesis rather than real-time visualization. Future work may explore adaptive density scheduling to balance efficiency and quality based on specific application needs.}

\subsection{Ablation Study}

\begin{table}[ht]
    \centering
    \caption{\hl{Ablation study results for the three key parameters in dynamic density control}}
    \label{tab:ablation_results}
    \begin{tabular}{c|*{3}{wc{0.7cm}}|*{3}{wc{0.7cm}}}
        \toprule
        \multirow{2}{*}{\textbf{Parameter}} & \multicolumn{3}{c}{\textbf{First Phase}} & \multicolumn{3}{c}{\textbf{Second Phase}} \\
        & \textbf{PSNR}↑ & \textbf{SSIM}↑ & \textbf{LPIPS}↓ & \textbf{PSNR}↑ & \textbf{SSIM}↑ & \textbf{LPIPS}↓ \\
        \midrule
        \( P_{di_{\text{def}}} \)  & 14.21  & 0.533  & 0.633  & -  & -  & - \\
        \( P_{di_{\text{NC}}} \)  & \textbf{18.68} & \textbf{0.764} & \textbf{0.421} & \textbf{28.71} & \textbf{0.901} & \textbf{0.273} \\
        \( P_{di_{\text{Inv}}} \)  & 12.23  & 0.566  & 0.674  & -  & -  & - \\
        \midrule
        \( P_{oi_{\text{def}}} \)  & 16.32  & 0.612  & 0.557  & 24.22  & 0.812  & 0.362 \\
        \( P_{oi_{\text{NC}}} \) & \textbf{18.68} & \textbf{0.764} & \textbf{0.421} & \textbf{28.71} & \textbf{0.901} & \textbf{0.273} \\
        \( P_{oi_{\text{Inv}}} \) & 15.51  & 0.588  & 0.593  & 20.58  & 0.765  & 0.501 \\
        \midrule
        \( \tau_{pos_{\text{def}}} \) & 16.25  & 0.611  & 0.526  & 24.98  & 0.721  & 0.341 \\
        \( \tau_{pos_{\text{NC}}} \)  & \textbf{18.68} & \textbf{0.764} & \textbf{0.421} & \textbf{28.71} & \textbf{0.901} & \textbf{0.273} \\
        \( \tau_{pos_{\text{Inv}}} \)  & 15.50  & 0.588  & 0.568  & 23.83  & 0.694  & 0.369 \\
        \bottomrule
    \end{tabular}
\end{table}

\begin{figure}
    \centering
    \includegraphics[width=0.95\linewidth]{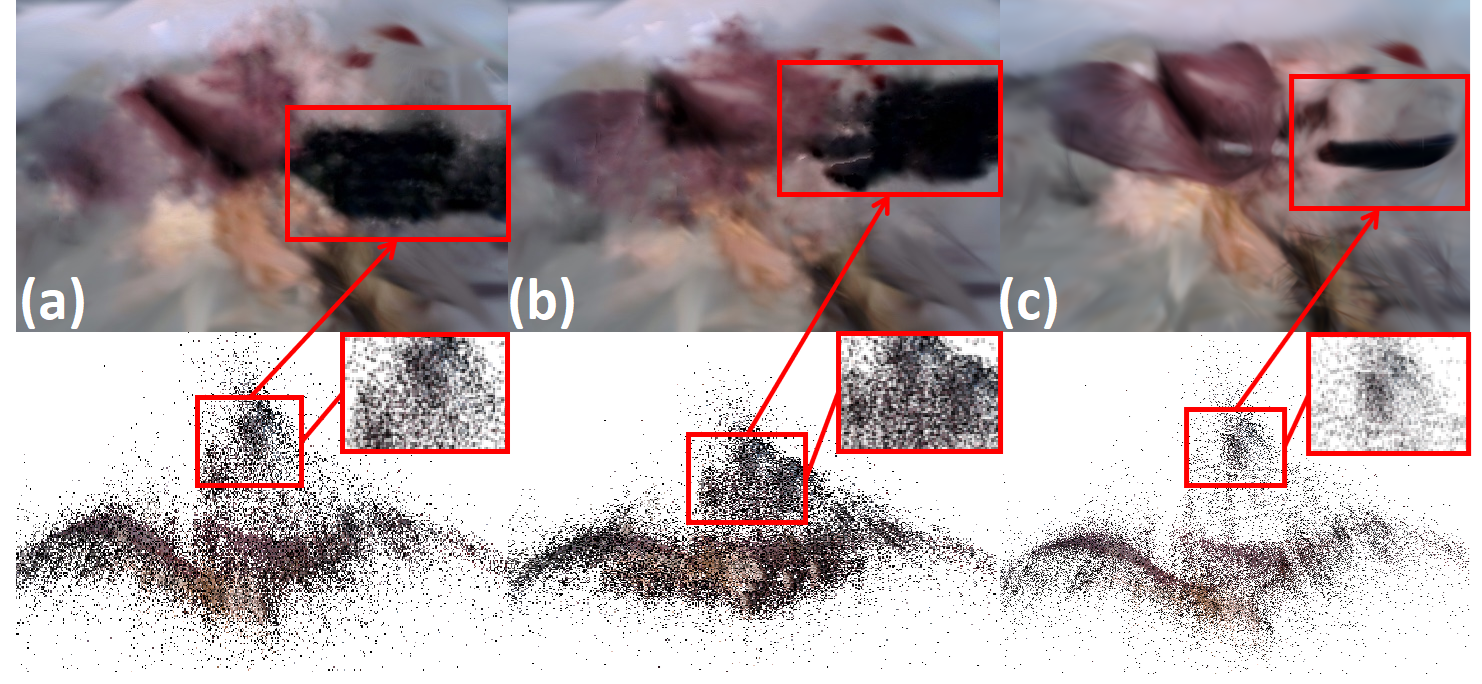}
    \caption{Ablation study on Dynamic Density Control, indicating differences in Gaussian quality at 6k training iterations, (a) Kerbl et al. \mbox{\cite{kerbl20233d}}, (b) Zhang et al. \mbox{\cite{zhang2024pixel}} (c) Ours.}
    \label{fig:ablation2}
\end{figure}

\begin{figure}[b]
    \centering
    \includegraphics[width=1\linewidth]{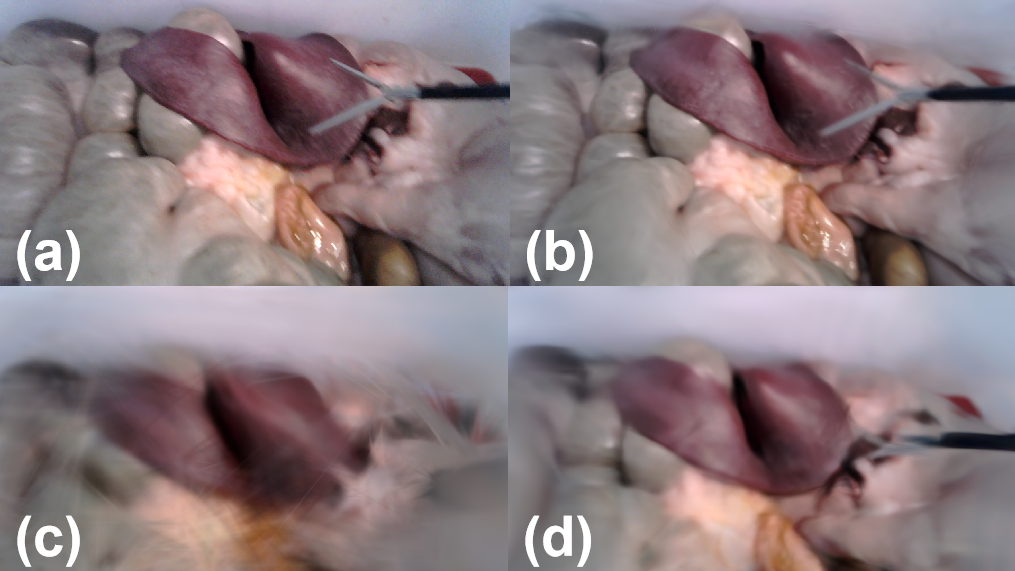}
    \caption{(a) GT image, (b) results with both Dynamic Density Control and Uniform Motion Rendering applied, (c) without Dynamic Density Control, (d) without Uniform Motion Rendering.}
    \label{fig:ablation}
\end{figure}

\begin{table}[ht]
\centering
\caption{Ablation Study Results}
\label{abresult}
\begin{tabular}{ccc|cccc}
\\ \toprule
\multirow{2}{*}{Metric} & \multirow{2}{*}{SH} & \multirow{2}{*}{Compensation} & \multicolumn{4}{c}{Epochs} \\

 &  &  & 4k & 8k & 12k & 20k \\
\midrule
\multirow{4}{*}{PSNR$\uparrow$} & \ding{55} & \ding{55} & 15.21 & 16.37 & 19.02 & 20.63 \\
 & \ding{51} & \ding{55} & 15.34 & 16.77 & 20.64 & 22.53\\
 & \ding{55} & \ding{51} & 16.73 & 18.69 & 20.66 & 22.81\\
 & \ding{51} & \ding{51} & \textbf{16.82} & \textbf{18.88} & \textbf{21.54} & \textbf{23.78}\\
\bottomrule
\end{tabular}
\end{table}

\subsubsection{Dynamic Density Control and Uniform Motion Rendering}
In this section, we evaluate the impact of Dynamic Density Control and Uniform Motion Rendering on the training process. First, we compared the effects of dynamic density control with other density control strategies on training. As illustrated in Figure \ref{fig:ablation2} during the initial training phase, both (a) and (b) increase point cloud density but fail to correct errors caused by poor initialization, leading to the over-generation of Gaussian representations for the instruments. In contrast, (c) effectively manages density, preventing the spread of erroneous points and ensuring accurate Gaussian representation of the instruments (PSNR: (a) 15.76, (b) 14.21, (c) 17.33). Additionally, we visualize the effects of different rendering strategies on training. As shown in Figure \ref{fig:ablation}, the absence of Dynamic Density Control (c) leads to extremely poor rendering quality, while the lack of Uniform Motion Rendering (d) results in inaccurate rendering of the surgical instruments (PSNR: (b) 28.31, (c) 15.42, (d) 22.26). Without Dynamic Density Control, our model had nearly an 80\% chance of failing to complete the training in our preliminary experimentation when tuning the SfM settings, underscoring the importance of this method in our study.
\hl{To further analyze the role of Dynamic Density Control, we evaluate the influence of key parameters: densification interval ($P_{di}$), opacity reset interval ($P_{oi}$), and positional gradient threshold ($\tau_{pos}$). Each parameter is assessed under three conditions: the default 3DGS setting ($_{\text{def}}$), our proposed NeeCo strategy ($_{\text{NC}}$), and an inverse setting ($_{\text{Inv}}$) where parameter adjustments follow an opposite trend. The results in Table \mbox{\ref{tab:ablation_results}} highlight the necessity of adaptive control across different training phases.

$P_{di}$ affects Gaussian point generation, influencing scene density and error propagation. In phase one, premature densification amplifies initialization errors, leading to unstable learning, as observed in the inverse setting ($P_{di_{\text{Inv}}}$), which results in PSNR dropping to 12.23. In contrast, restricting early densification ($P_{di_{\text{NC}}}$) stabilizes training and improves PSNR by 31.5\% compared to the default setting. Once erroneous Gaussians are corrected in phase two, increasing $P_{di}$ facilitates better scene representation, achieving a PSNR of 28.71 and SSIM of 0.901, whereas both $P_{di_{\text{def}}}$ and $P_{di_{\text{Inv}}}$ fail to converge due to accumulated early-stage errors.

$P_{oi}$ controls the removal of low-opacity Gaussians, balancing noise suppression and scene refinement. In phase one, delaying opacity resets prevents premature removal of useful Gaussians, improving PSNR by 14.5\% over the default setting. Conversely, in the inverse setting ($P_{oi_{\text{Inv}}}$), frequent removals cause excessive sparsity, reducing PSNR to 15.51. In phase two, frequent resets refine geometric detail, where $P_{oi_{\text{NC}}}$ achieves SSIM 0.901, while insufficient resets in $P_{oi_{\text{Inv}}}$ degrade performance to PSNR 20.58.

$\tau_{pos}$ regulates Gaussian movement constraints. A higher threshold in phase one stabilizes training, with $\tau_{pos_{\text{NC}}}$ achieving PSNR 18.68, outperforming $\tau_{pos_{\text{def}}}$ (PSNR 16.25). In phase two, lowering $\tau_{pos}$ enables finer positional refinement, leading to a 15\% PSNR improvement over the default setting, while an overly low threshold in $\tau_{pos_{\text{Inv}}}$ causes unstable convergence (PSNR 23.83).

These findings validate that static hyperparameter tuning in 3DGS is insufficient for dynamic surgical instrument reconstruction. The NeeCo strategy effectively stabilizes early training, prevents error accumulation, and enables successful convergence even in previously unstable models. Additionally, our ablation study results further support the necessity of Dynamic Density Control in achieving robust 3D Gaussian-based scene representation.}

\subsubsection{Dynamic SH Update and Camera Pose Compensation}
We compared the impact of Dynamic SH function updates and camera pose compensation on training. As shown in Table \ref{abresult}, the model without any implementation yielded the poorest rendering quality. The model with only SH updates showed minial improvement in the early stages of training (4k-8k epochs) but demonstrated noticeable gains in the later stages (12k-20k epochs), as the early SH updates focused on lower orders, allowing the model to better address erroneous initializations. The model with Dynamic Camera Compensation exhibited consistent PSNR improvements throughout the training process. The best results were achieved when both methods were applied together.

\section{Discussion}

 \hl{Recent research has focused on leveraging 3DGS to model surgical background deformation by encoding time-dependent deformation fields \mbox{\cite{yang2024deform3dgs, huang2024endo, zhao2024hfgs}}. These methods primarily focus on capturing and reproducing previously observed tissue deformations by tracking temporal changes in continuous video sequences. While effective in recreating background dynamics, they rely heavily on sequential image data and cannot predict unseen deformations or generate structured instrument labels. In contrast, NeeCo takes a fundamentally different approach by focusing on instrument-centric scene reconstruction and dataset generation. This is achieved by learning instrument poses rather then sequential video frames. Unlike tissue-focused reconstruction methods, our approach is not constrained by temporally continuous frames, allowing for greater flexibility in dataset expansion. 

While NeeCo can outperform SOTA methods in both ground truth geeneration and image synthesis, it currently relies on 7-DoF kinematic data obtained EM tracking system and a jaw angle sensor. This setup ensures accurate ground truth for instrument motion, facilitating the synthesis of high-fidelity training data. However, in real-world surgical scenarios, such precise kinematic tracking systems are not always available, which may limit the direct applicability of the approach. Recent advancements in deep learning-based pose estimation have demonstrated promising results in estimating instrument motion without relying on external tracking devices \mbox{\cite{jiang2023development}}. Incorporating such techniques into our framework could enhance its applicability by enabling dataset generation using standard endoscopic video recordings.

NeeCo assumes a static tissue background, which may not fully represent real surgical conditions where soft tissue deformations occur due to instrument interaction. Where other methods focus on reconstucting previously seen tissue deformations, we assume here that the tools do not contact the background tissue. In order to predict tissue deformations following a fully unseen movement sequence the method would need to incorporate dynamic tissue modeling, such as learnable deformation priors or physics-based tissue deformation models. Dynamic models could further improve the realism of synthetic datasets and broaden their application to more complex surgical environments, but are as yet unsolved and fall outside the scope of this work.

During our experiments, we observed that NeeCo encounters challenges in specific scenarios, leading to training instability or degraded rendering quality. First, a limited training range of observed poses a significant challenge for reconstructing a complete 3D representation of the tool. Since NeeCo relies on unordered discrete images rather than sequential time-encoded inputs, insufficient tool motion range results in incomplete understanding of the tooland overfitting to the available perspectives, limiting the model's ability to reconstruct the full geometry of the instrument in all states. Consequently, unseen viewpoints in the test phase exhibit structural instability or incomplete reconstructions particularly in the tool jaw. Second, rapid motion in the dataset capture causes significant motion blur in the images. As a result, the network struggles to establish consistent Gaussian point correspondences across different views, disrupting global geometric consistency. In high-speed motion scenarios, the rendered results exhibit noticeable blurriness, leading to reduced sharpness and structural consistency in the synthesized images.}

\section{Conclusion}
This paper presents a novel pipeline for generating surgical instrument deformation images, which, compared to existing methods, contributes to creating realistic and diverse surgical image datasets. Our approach introduces dynamic 3D Gaussian models to represent the deformation of instruments in dynamic surgical scenes and employs a dynamic density control strategy to address the challenges posed by poor camera poses in real-world datasets, which often hinder training.

Additionally, our method can generate annotation files, addressing the significant challenge of the lack of annotated data in medical imaging datasets. Our experiments demonstrate promising results, outperforming recent work and achieving object detection and segmentation performances that closely resemble those of models trained on GT imaging. Moreover, the datasets generated using our method's capability to render new deformations and viewpoints further surpass the performance of models trained solely on GT imaging. 

However, our method has limitations. It struggles to capture background tissue deformations accurately, particularly when such deformations follow a specific temporal or operational sequence. Our future work will address the challenge of predicting soft tissue deformations from unordered input and improving the recovery of these deformations.

\end{document}